\documentclass[runningheads]{llncs}
\usepackage[T1]{fontenc}
\usepackage{graphicx}
\usepackage{amssymb}
\usepackage{amsmath}
\usepackage{siunitx}
\usepackage{bm}
\usepackage{hyperref}
\usepackage{color}

\begin{document}
\title{Differentiable Score-Based Likelihoods: Learning CT Motion Compensation From Clean Images}
\titlerunning{Learning CT Motion Compensation From Clean Images}
% If the paper title is too long for the running head, you can set
% an abbreviated paper title here
%

% \author{*****\\
% *****\\
% *****
% }

\author{Mareike~Thies\inst{1} \and
Noah~Maul\inst{1} \and
Siyuan~Mei\inst{1} \and
Laura~Pfaff\inst{1} \and
Nastassia~Vysotskaya\inst{1} \and
Mingxuan~Gu\inst{1} \and
Jonas~Utz\inst{1} \and
Dennis~Possart\inst{1,2} \and
Lukas~Folle\inst{1} \and
Fabian~Wagner\inst{1} \and
Andreas~Maier\inst{1}}

\authorrunning{M. Thies et al.}
% First names are abbreviated in the running head.
% If there are more than two authors, 'et al.' is used.
%
% \institute{***** \and
% *****\\
% \email{*****}}

\institute{Friedrich-Alexander-Universit\"at Erlangen-N\"urnberg, Erlangen, Germany \and
Fraunhofer Institute for Ceramic Technologies and Systems IKTS, Forchheim, Germany\\
\email{mareike.thies@fau.de}}

\maketitle              % typeset the header of the contribution
\begin{abstract}
Motion artifacts can compromise the diagnostic value of computed tomography (CT) images. Motion correction approaches require a per-scan estimation of patient-specific motion patterns. In this work, we train a score-based model to act as a probability density estimator for clean head CT images. Given the trained model, we quantify the deviation of a given motion-affected CT image from the ideal distribution through likelihood computation. We demonstrate that the likelihood can be utilized as a surrogate metric for motion artifact severity in the CT image facilitating the application of an iterative, gradient-based motion compensation algorithm. By optimizing the underlying motion parameters to maximize likelihood, our method effectively reduces motion artifacts, bringing the image closer to the distribution of motion-free scans. Our approach achieves comparable performance to state-of-the-art methods while eliminating the need for a representative data set of motion-affected samples. This is particularly advantageous in real-world applications, where patient motion patterns may exhibit unforeseen variability, ensuring robustness without implicit assumptions about recoverable motion types.     

\end{abstract}

\section{Introduction}
Spotting motion artifacts in CT images is an easy task. Even an inexperienced CT reader can differentiate an artifact-free from a motion-affected image. As observers of the images, we have an implicit understanding of a ``good'' CT image. Therefore, we can recognize deviations from that state as, e.g., introduced through motion. In contrast, neural networks usually need to be presented with examples of all possible states, i.e., motion-free and motion-affected images to grade the intensity of motion artifacts in a CT image \cite{preuhs2020,huang2022,thies2024}. During training, this requires a representative labeled data set including images of all possible expected motion states and their respective motion score. %These data sets are commonly simulated from motion-free scans by applying motion trajectories to the CT geometry during reconstruction. This simulation introduces assumptions about the expected type of motion patterns. 

We aim to emulate the human observer with a network that identifies motion artifacts after training on clean images only. During the training process, the network learns to model the distribution of clean head CT images. Motion artifacts can then be quantified by the deviation of a sample from this distribution. In this work, we evaluate the likelihood of a given, potentially motion-affected CT image under the distribution of motion-free images represented by the trained model. Not only does this technique allow for accurate motion artifact quantification, it additionally yields enough information for gradient-based optimization of the underlying motion patterns. This allows for motion artifact compensation simply by pulling the image closer to the distribution of motion-free samples. 
%We demonstrate that this measure not only quantifies motion artifacts accurately, but even yields enough information for gradient-based optimization of the underlying motion pattern to compensate for these artifacts by pulling the image closer to the distribution of motion-free samples.

\textbf{Motion compensation:} Motion artifacts occur when the patient moves during the acquisition of a CT scan. This introduces a mismatch between the measured CT geometry and the one assumed during reconstruction resulting in streaks, double edges, and blurring in the final image. Most works for motion artifact reduction estimate the underlying motion pattern of a specific scan followed by a motion-compensated reconstruction. The motion estimate can be obtained from external measurements or attached markers \cite{maier2020,choi2013}. Alternatively, existing approaches optimize consistency conditions on the projection data \cite{yu2007,berger2017,aichert2015} or evaluate the quality of intermediate reconstructions in the image domain \cite{kingston2011,sisniega2017,preuhs2020,huang2022}. The key disadvantage of consistency-based solutions is their requirement for non-truncated projection images. Consequently, this work focuses on the estimation of rigid head motion by iteratively reconstructing the data and optimizing the quality of these intermediate reconstructions.

\textbf{Out-of-distribution and anomaly detection:} Anomaly and out-of-dis\-tri\-bution (OOD) detection are concerned with identifying samples that differ from a known distribution, mostly the training distribution \cite{mueller2023}. These approaches often reconstruct the data through some sort of bottleneck which captures in-distribution characteristics, but fails to reconstruct OOD features \cite{denouden2018,zhou2022,graham2023}. Alternatively, generative networks have been employed as density estimators for the known distribution for which likelihoods can be evaluated as a proxy for the similarity of a new sample to the training data distribution \cite{ren2019,xiao2020,zisselman2020}. Such likelihood-based methods have been proposed using normali\-zing flow models, energy-based models, or variational autoencoders. However, for score-based diffusion models, most existing work on OOD or anomaly detection is focused around reconstruction \cite{wolleb2022,linmans2024,choi2024} rather than likelihood evaluation \cite{goodier2023}. 

\textbf{Contributions:} In this paper, we present a motion correction method that is solely trained on a data set of clean CT images. The method thus circumvents the construction of data sets with motion artifacts that represent all possible motion states. Our contributions are (1) the grading of an image's motion-affectedness via the exact likelihood of a diffusion model trained on motion-free images, (2) the application of a neural ordinary differential equation (ODE) solver \cite{chen2018} inside the likelihood function to calculate the gradient of the likelihood value with respect to the image, and (3) the application of this differentiable likelihood function as an objective for state-of-the-art, gradient-based motion compensation.  

\section{Methods}
\subsection{Score-Based Diffusion Models}
Score-based diffusion models approximate a data distribution by estimating gradients of the corresponding probability density \cite{song2021}. Based on a diffusion process, they transform the data distribution $p_{\mathrm{data}}$ at time $t=0$ into an easily tractable prior distribution $p_{\mathrm{prior}}$ at time $t=T$. For image generation, this process can be reversed in time and samples from the prior distribution are transformed into samples from the data distribution by solving a stochastic differential equation (SDE) backward in time. Interestingly, for each such SDE, there exists a corresponding ordinary differential equation (ODE) which is deterministic and shares the same marginal distributions $p_t(\mathbf{x})$ at any time point $t$
\begin{equation}
    \label{eq:ode}
    \mathrm{d}\mathbf{x} = \left[ \mathbf{f}(\mathbf{x}, t) - \frac{1}{2}g(t)^2 \mathbf{\nabla_x} \log p_t(\mathbf{x}) \right]\mathrm{d}t \enspace .
\end{equation}
This ODE is frequently referred to as probability flow ODE. Using a variance exploding variant, the drift coefficient is defined as $\mathbf{f}(\mathbf{x}, t) = \mathbf{0}$ and the diffusion coefficient is $g(t) = \sigma_t \sqrt{2 \log \frac{\sigma_{\mathrm{max}}}{\sigma_{\mathrm{min}}}}$ where $\sigma_t\in[\sigma_{\mathrm{min}}, \sigma_{\mathrm{max}}]$ denotes the noise level at time point $t$. The gradient of the log probability density function $\mathbf{\nabla_x} \log p_t(\mathbf{x})$ can be approximated by a time-conditional score network $\mathbf{s}_{\mathbf{\theta}}$ such that $\mathbf{s}_{\mathbf{\theta}}(\mathbf{x}, t) \approx \mathbf{\nabla_x} \log p_t(\mathbf{x})$. We refer the reader to \cite{song2021} for details on the training process of $\mathbf{s}_{\mathbf{\theta}}$.

\subsection{Likelihood Computation}
Given a trained score network, we can generate a sample $\mathbf{x}(0)$ from the data distribution by integrating Equation~\ref{eq:ode} backward in time starting from an initial sample $\mathbf{x}(T)$ from the prior distribution 
\begin{equation}
    \mathbf{x}(0) = \mathbf{x}(T) + \int_{T}^{0} \underbrace{\left[ \mathbf{f}(\mathbf{x}, t) - \frac{1}{2}g(t)^2 \mathbf{s}_{\mathbf{\theta}}(\mathbf{x}, t) \right]}_{\tilde{\mathbf{f}}_{\mathbf{\theta}}(\mathbf{x}, t)}\mathrm{d}t \enspace .
\end{equation}
Now, we can apply the continuous change of variables formula to this implicit transformation between prior and data distribution to compute the log likelihood of $\mathbf{x}(0)$ \cite{song2021,chen2018,kothe2023}
\begin{equation}
    \label{eq:likelihood}
    \log p(\mathbf{x}(0)) = \log p(\mathbf{x}(T)) + \int_{0}^{T} \left[ \mathrm{tr}\left( \mathbf{\nabla} \tilde{\mathbf{f}}_{\mathbf{\theta}}(\mathbf{x}, t) \right) \right] \mathrm{d}t \enspace .
\end{equation}
Since the trace of the Jacobian $\mathrm{tr}(\mathbf{\nabla} \tilde{\mathbf{f}}_{\mathbf{\theta}}(\mathbf{x}, t))$ is computationally expensive, we follow other existing approaches \cite{song2021,grathwohl2018} and approximate it with the Hutchinson trace estimator \cite{hutchinson1989}
\begin{equation}
    \mathrm{tr}\left( \mathbf{\nabla} \tilde{\mathbf{f}}_{\mathbf{\theta}}(\mathbf{x}, t) \right) = \mathbb{E}_{\epsilon} \left[ \epsilon^T \mathbf{\nabla} \tilde{\mathbf{f}}_{\mathbf{\theta}}(\mathbf{x}, t) \epsilon \right] \approx \frac{1}{M} \sum_{m=1}^M \epsilon_m^T \mathbf{\nabla} \tilde{\mathbf{f}}_{\mathbf{\theta}}(\mathbf{x}, t) \epsilon_m \enspace ,
\end{equation}
which can be evaluated efficiently using automatic differentiation for a limited number $M$ of random vectors $\epsilon_m$ drawn from the Rademacher distribution. 
Ultimately, we can evaluate the likelihood in Equation~\ref{eq:likelihood} for any given data sample $\mathbf{x}(0)$ by solving an ODE forward in time. In \cite{song2021}, Song et al.\ use an adaptive step size Runge-Kutta solver of order five from the \textit{scipy} library and $M=5$ realizations of random vectors $\epsilon_m$ to solve the ODE with high accuracy. 
% todo: what about the bpd scaling?

\subsection{Neural ODEs and Likelihood Target Function}
 In this work, we use Equation~\ref{eq:likelihood} as the objective function in a gradient-based optimization problem for CT motion compensation. This entails two implications: (1) We can not simply utilize any ODE solver, but we require differentiability through the solver to obtain the gradient of the log likelihood $\log p(\mathbf{x}(0))$ with respect to the image $\mathbf{x}(0)$, and (2) we do not necessarily need to solve Equation~\ref{eq:likelihood} to the highest possible precision as long as the loss landscape remains well-behaved. 
 
 To address these requirements, we substitute the conventional ODE solver with a neural ODE solver\footnote{\url{https://github.com/rtqichen/torchdiffeq}} to ensure differentiability. Instead of differentiating through the internals of the solver by tracking the computations of the forward pass, it solves the so-called adjoint ODE for gradient computation \cite{chen2018}. This ensures differentiable behavior at low memory cost since intermediate results do not need to be stored. To reduce the computation time, we configure a fixed step size Runge-Kutta solver of order four with ten steps for the forward ODE and 20 steps for the gradient calculation via the adjoint ODE. Moreover, for each evaluation of Equation~\ref{eq:likelihood}, we sample one realization of $\epsilon_m$ and keep it constant for the gradient computation. Repeated evaluations of the target function use different random vectors $\epsilon_m$.  
 
\begin{figure}[t]
    \centering
    \includegraphics[width=\textwidth]{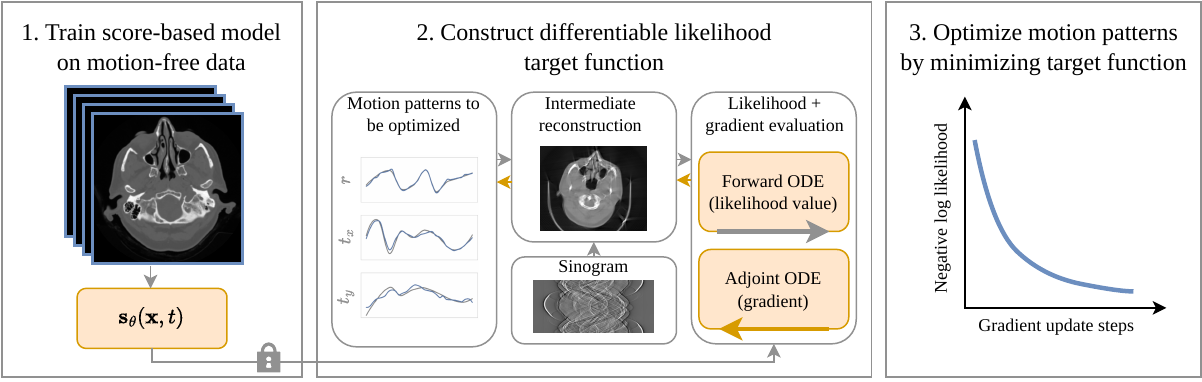}
    \caption{Overview of the proposed pipeline. After training on motion-free images, the score model $\mathbf{s}_{\mathbf{\theta}}(\mathbf{x}, t)$ is used inside the likelihood target function. For each step of the gradient descent optimizer, we perform an intermediate reconstruction and evaluate the gradient of the likelihood function to pull the image closer to the distribution of clean images seen during training.}
    \label{fig:pipeline}
\end{figure}
\subsection{Motion Compensation}
Figure~\ref{fig:pipeline} summarizes the proposed algorithm using the differentiable likelihood function as the target function for motion compensation. First, a score network is fitted to the distribution of motion-free images which is subsequently applied inside the likelihood target function in Equation~\ref{eq:likelihood}. For motion compensation, we estimate a set of splines describing the motion trajectories. Each spline parameterizes the values for one of the three rigid motion parameters (translations $t_x$ and $t_y$ and rotation $r$) throughout the scan with 30 evenly distributed nodes yielding 90 free parameters $\bm{\gamma}^{(n)}$ in total for the optimization problem at step $n$. In each iteration of the optimization algorithm, we (1) apply the current state of motion parameters to the fan-beam geometry, (2) reconstruct the sinogram based on this trajectory to obtain an intermediate, possibly motion-affected image $\mathbf{x}^{(n)}(0)$, and (3) feed this image into the previously introduced likelihood function to grade its motion-affectedness in terms of a negative log likelihood value. Next, the solution to the adjoint ODE yields the gradient of the likelihood value with respect to the intermediate reconstructed image $\frac{\mathrm{d} \log p(\mathbf{x}^{(n)}(0))}{\mathrm{d} \mathbf{x}^{(n)}(0)}$ which is subsequently propagated into the free motion parameters using the Jacobian of the fan-beam reconstruction operator $\frac{\mathrm{d} \mathbf{x}^{(n)}(0)}{\mathrm{d} \bm{\gamma}^{(n)}}$ \cite{thies2023,thies2024}
\begin{equation}
    \frac{\mathrm{d} \log p(\mathbf{x}^{(n)}(0))}{\mathrm{d} \bm{\gamma}^{(n)}} = \frac{\mathrm{d} \log p(\mathbf{x}^{(n)}(0))}{\mathrm{d} \mathbf{x}^{(n)}(0)} \cdot \frac{\mathrm{d} \mathbf{x}^{(n)}(0)}{\mathrm{d} \bm{\gamma}^{(n)}} \enspace .
\end{equation}
Given the target function's gradient with respect to the free parameters, the spline-based motion patterns are optimized using basic gradient descent 
\begin{equation}
    \label{eq:gradient_descent}
    \bm{\gamma}^{(n + 1)} = \bm{\gamma}^{(n)} + r^{(n)} \cdot \frac{\mathrm{d} \log p(\mathbf{x}^{(n)}(0))}{\mathrm{d} \bm{\gamma}^{(n)}}
\end{equation}
with an initial estimate $\bm{\gamma}^{(0)} = \mathbf{0}$ and an exponentially decaying step size $r^{(n)} = r_0 \cdot q^n$. 40 iterations of Equation~\ref{eq:gradient_descent} are executed with an initial step size of $r_0 = 100$ and a decay factor of $q=0.97$. 

\section{Experiments and Results}
Our experiments are performed retrospectively on publicly available head CT scans from the CQ500 data set published under CC BY-NC-SA license \cite{chilamkurthy2018}. All terms of use in the end-user license agreement were followed strictly. The score network is trained on slices from 200 subjects and evaluated on slices from 40 subjects. The motion compensation experiments are performed on slices from another 40 disjunct subjects. For the score network, we use a reduced-parameter version of the NCSN\texttt{++} architecture proposed in \cite{song2021}. Exemplary generated head CT samples from the trained model can be found in the supplementary material. 

Motion compensation is performed on individual slices. A sinogram is simulated for each slice by forward projecting it onto a trajectory with 360 projections on a full circle, a source-to-isocenter distance of \qty{785}{\milli\meter}, a source-to-detector distance of \qty{1200}{\milli\meter}, and a detector with 700 elements at spacing \qty{0.64}{\milli\meter}. Reconstructions are computed on a grid of $256 \times 256$ pixels with an isotropic spacing of \qty{1}{\milli\meter}. The initial motion-affected state is obtained by perturbing the ground-truth reconstruction trajectory with a spline-based motion pattern with ten nodes and a maximal amplitude of \qty{5}{\milli\meter} for translations $t_x$ and $t_y$ as well as \ang{5} for rotation $r$. The likelihood-based target function is compared to two alternative objectives. Importantly, both reference methods require substantially more knowledge about the problem than our proposed method. First, as an upper performance bound, motion parameters are optimized by computing the mean-squared error (MSE) to the ground-truth, motion-free reconstruction. This represents an ideal target function, but requires knowledge of the motion-free image at optimization time, making it impractical for almost all applications. Second, following existing approaches \cite{thies2023,preuhs2020,huang2022}, we train a network on a paired data set of motion-affected and motion-free images to predict the structural si\-mi\-larity index measure (SSIM) given only the motion-affected image. This trained network is referred to as autofocus target function in the following. Notably, in contrast to the proposed target function, the autofocus network has seen examples of motion-affected images during training.   

\begin{figure}[t]
    \centering
    \includegraphics[trim={2.0cm 1.2cm 3cm 1cm},clip,width=\textwidth]{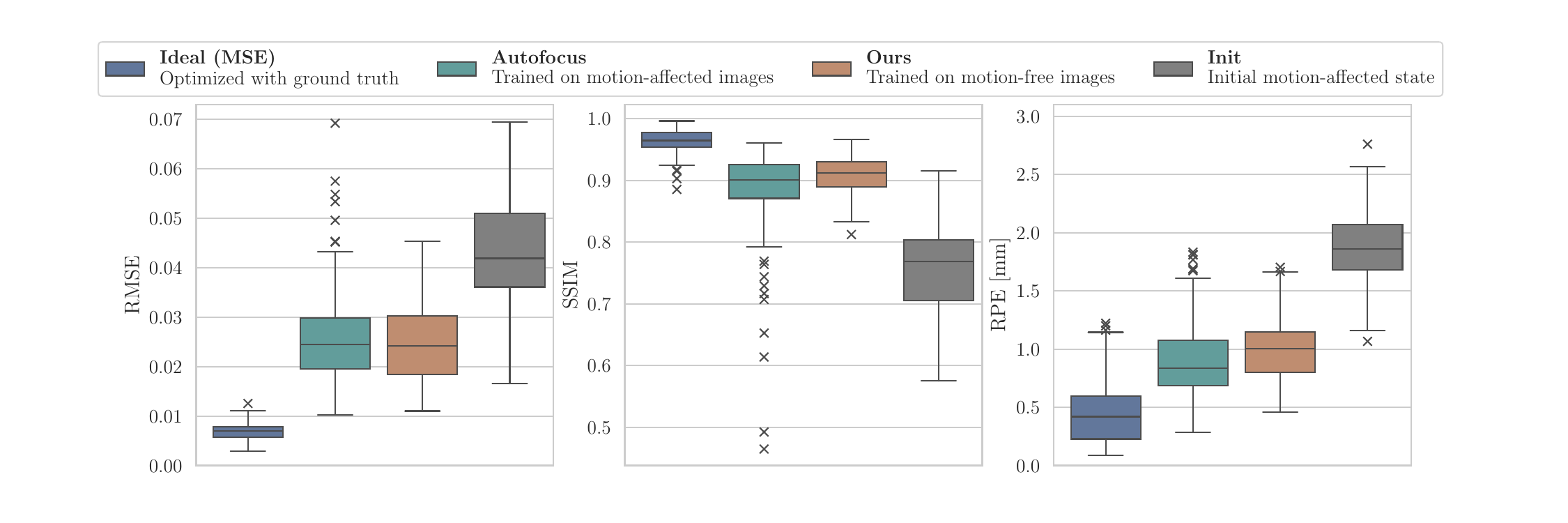}
    \caption{Comparison of target functions using root mean squared error (RMSE) ($\downarrow$) and SSIM ($\uparrow$) of the motion-compensated images as well as reprojection error (RPE) ($\downarrow$). The proposed method achieves similar results as the autofocus method despite never having seen any motion-affected images during training. The MSE target function can be considered an upper performance bound since it requires knowledge of the motion-free image at optimization time which is unrealistic in a clinical workflow.}
    \label{fig:results}
\end{figure}
\begin{table}
    \centering
    \caption{Average quantitative values for all investigated metrics comparing the proposed method to the autofocus target function and the initial motion-affected reconstruction. The ideal MSE target function is excluded since it requires knowledge of the motion-free ground truth at optimization time and is therefore impractical. The best value is highlighted in bold font.}
    \label{tab:results}
    \begin{tabular}{@{}lcccccc}
    \hline
        & RMSE [$\times0.1$] ($\downarrow$) & SSIM ($\uparrow$) & RPE ($\downarrow$) & MAE $t_x$ ($\downarrow$) & MAE $t_y$ ($\downarrow$) & MAE $r$ ($\downarrow$) \\
    \hline
        Init        & 0.45 & 0.74 & 1.98 & 1.01 & 1.02 & 0.98\\
        Autofocus   & 0.26 & 0.89 & \textbf{0.94} & 0.75 & 0.77 & \textbf{0.44}\\
        Ours        & \textbf{0.25} & \textbf{0.91} & 0.99 & \textbf{0.71} & \textbf{0.72} & 0.51\\
        %\hline
        %Ideal (MSE) & 0.07 & 0.97 & 0.45 & 0.55 & 0.54 & 0.35\\
    \hline
    \end{tabular}
\end{table}

\begin{figure}[t]
    \centering
    \includegraphics[width=\textwidth]{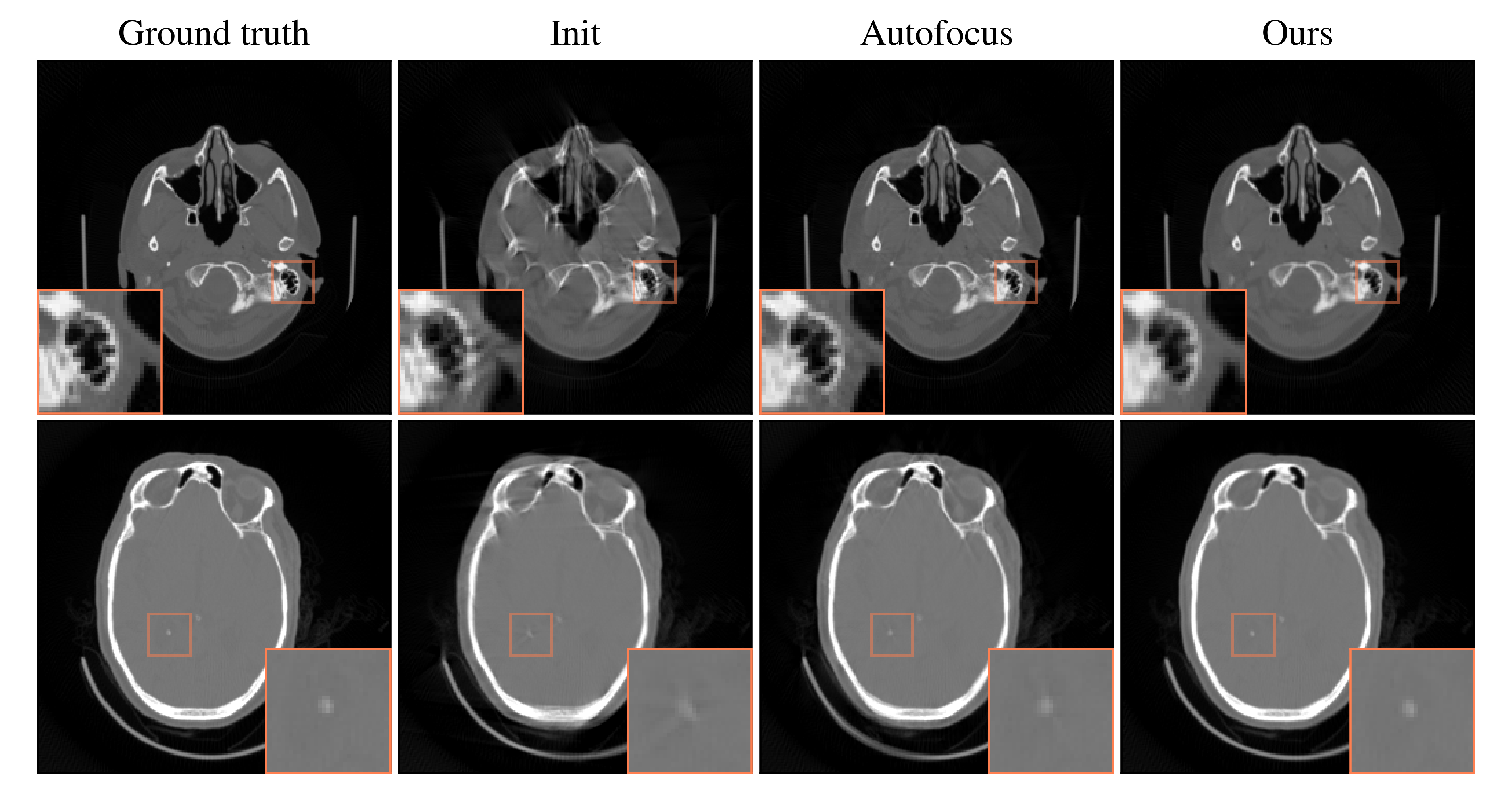}
    \caption{Two example slices before and after motion compensation with autofocus and likelihood objective. A $\times3$ zoom of a region of interest is shown in the orange frame.}
    \label{fig:recos}
\end{figure}

Figure~\ref{fig:results} illustrates the motion compensation performance of the investigated target functions. Overall, all methods improve upon the initial motion-affected state. The MSE-based optimization with known ground truth performs best. This is to be expected and can be regarded as the upper performance bound of the motion compensation pipeline given an ideal target function. The autofocus target function and the likelihood-based objective proposed in this work perform similarly concerning the RMSE and SSIM of the motion-compensated images. The autofocus method results in a slightly lower RPE, but higher variance and more negative outliers. These findings are confirmed in Table~\ref{tab:results}. Additionally, we compute the mean absolute error (MAE) for each of the three motion parameters ($t_x$, $t_y$, and $r$). While the likelihood objective yields a higher MAE for rotation, it recovers translations slightly better than the autofocus approach. Figure~\ref{fig:recos} shows two example slices before and after motion compensation with the autofocus and likelihood target function. Particularly in the enlarged regions of interest, a clear improvement over the initial motion-affected state can be observed with fewer streaks and less blur. The compensated image from the likelihood objective appears slightly smoother than the one from the autofocus objective, but both successfully restore even fine details such as the bright spot in the bottom example.     

\section{Discussion}
The proposed method performs on par with the autofocus approach despite the fact that the autofocus network was trained on motion-affected examples of different motion patterns and motion amplitudes. In contrast, the proposed motion compensation method only needs motion-free images. Practically, this is an enormous difference since multiple data sets of clean CT images are publicly available. On the other hand, creating a paired data set of motion-free and motion-affected data as needed by the autofocus-type approaches is not straightforward. The simulation of representative motion patterns that are applied to the projection geometry for motion perturbation is complex and usually based on more or less realistic assumptions about the actual motion of patients in a scanner. As a result, specific motion patterns occurring in practice might not be covered. The likelihood-based approach is free of such assumptions. We acknowledge that evaluating the proposed target function and its gradient is considerably slower than the other objectives because it needs to solve two ODEs per gradient update step which both depend on the score network. Hence, the forward and backward pass of the score network are evaluated multiple times for a single update of the motion parameters. This property is inherited from the diffusion process in score-based modeling which also makes sample generation slower compared to other generative models. However, speeding up the sampling is an active area of research and we hypothesize that such approaches could improve the speed of likelihood computation as well \cite{zheng2022}. Additionally, we would like to comment on the findings of existing papers which generally question the usability of likelihoods for OOD or anomaly detection since they seem to be systematically biased by the complexity of the data \cite{nalisnick2018,caterini2022}. Whereas our proposed target function is inspired by the ideas of likelihood-based OOD detection, we only perform a per-sample optimization for motion compensation. Consequently, we do not require a total separation of motion-free and motion-affected image distributions as long as the motion-free state of any given individual sample is graded better than a motion-affected state of the same sample.  

\section{Conclusion}
In this work, we extend the exact likelihood computation of score-based models, making it applicable as a target function in gradient-based optimization. Applied to CT motion compensation, it eliminates the need to simulate examples of motion-affected images for training. Since we only model the distribution of clean images, the same target function could be envisioned for other image restoration tasks without changes posing an exciting direction for future research. 

\begin{credits}
\subsubsection{\ackname} 
The research leading to these results has received funding from the European Research Council (ERC) under the European Union’s Horizon 2020 research and innovation program (ERC Grant No. 810316). The authors gratefully acknowledge the scientific support and HPC resources provided by the Erlangen National High Performance Computing Center of the Friedrich-Alexander-Universität Erlangen-Nürnberg. The hardware is funded by the German Research Foundation.

\subsubsection{\discintname}
The authors have no competing interests to declare that are relevant to the content of this article.
\end{credits}

\newpage
%
% ---- Bibliography ----
%
% BibTeX users should specify bibliography style 'splncs04'.
% References will then be sorted and formatted in the correct style.
%
\bibliographystyle{splncs04}
\bibliography{bibliography}

\end{document}